\documentclass[journal]{IEEEtran}

\usepackage{xcolor,soul,framed} 

\colorlet{shadecolor}{yellow}
\usepackage[pdftex]{graphicx}
\graphicspath{{../pdf/}{../jpeg/}}
\DeclareGraphicsExtensions{.pdf,.jpeg,.png}

\usepackage[cmex10]{amsmath}
\usepackage{array}
\usepackage{mdwmath}
\usepackage{mdwtab}
\usepackage{eqparbox}
\usepackage{url}

\usepackage{amssymb}
\usepackage{amsmath,amsfonts}
\usepackage{algorithmic}
\usepackage{algorithm}
\usepackage{array}
\usepackage[caption=false,font=normalsize,labelfont=sf,textfont=sf]{subfig}
\usepackage{textcomp}
\usepackage{stfloats}
\usepackage{url}
\usepackage{verbatim}
\usepackage{graphicx}
\usepackage{multirow}
\usepackage{color}
\usepackage{bbding}
\usepackage{wrapfig}
\usepackage{graphicx}
\usepackage{stfloats}

\newcommand{\eg}{\textit{e}.\textit{g}.}

\newcommand{\M}[1]{{#1}}
\newcommand{\MM}[1]{{#1}}

\usepackage[pdfstartview=XYZ,
bookmarks=true,
colorlinks=true,
linkcolor=blue,
urlcolor=blue,
citecolor=blue,
pdftex,
bookmarks=true,
linktocpage=true, 
hyperindex=true
]{hyperref}

\usepackage{orcidlink}

\begin{document}
\bstctlcite{IEEEexample:BSTcontrol}
    \title{DPNet: Dynamic Pooling Network for Tiny Object Detection}
  \author{
 
  Luqi Gong\orcidlink{0009-0000-8744-8630},~\IEEEmembership{Student Member,IEEE},
  Haotian Chen\orcidlink{0009-0004-7634-1968},
  Yikun Chen\orcidlink{0009-0004-4550-7689},~\IEEEmembership{Member,IEEE},
  Tianliang Yao\orcidlink{0009-0000-7063-3880},~\IEEEmembership{Student Member,IEEE},
  Chao Li\orcidlink{0000-0002-5343-1862},~\IEEEmembership{Member,IEEE},
  Shuai Zhao\orcidlink{0000-0002-5217-004X},~\IEEEmembership{Member,IEEE},
  Guangjie Han\orcidlink{0000-0002-6921-7369},~\IEEEmembership{Fellow,IEEE}
  
  \thanks{Manuscript received XXXXXX. (Corresponding author: Shuai Zhao; Chao Li).}
  \thanks{Luqi Gong is with the School of Computer Science, Beijing University of Posts and Telecommunications, Beijing, China, and also with the Research Center for Space Computing System, Zhejiang Lab, Hangzhou, China(e-mail:luqi@bupt.edu.cn, luqi@zhejianglab.org).}
  \thanks{Haotian Chen is with the SWJTU-LEEDS Joint School, Southwest Jiaotong University, Chengdu, China(email: cht@my.swjtu.edu.cn).}
  \thanks{Yikun Chen is with the Guangdong Zhiyun City construction Technology Co., LTD, Zhuhai, China(email: kenter1643@163.com).}%
  \thanks{Tianliang Yao is with the Department of Control Science and Engineering, College of Electronic and Information Engineering, Tongji University, Shanghai, China(email: yaotianliang@tongji.edu.cn).}
  \thanks{Chao Li is with the Research Center for Space Computing System, Zhejiang Lab, Hangzhou, China(email: lichao@zhejianglab.com).}
  \thanks{Shuai Zhao is with the School of Computer Science, Beijing University of Posts and Telecommunications, Beijing, China(email: zhaoshuaiby@bupt.edu.cn).}
  \thanks{Guangjie Han is with the Key Laboratory of Maritime Intelligent Network Information Technology, Ministry of Education, Hohai University, China (e-mail: hanguangjie@gmail.com).}
  \thanks{Luqi Gong and Haotian Chen contribute equally to the article.}
  \thanks{Copyright (c) 20xx IEEE. Personal use of this material is permitted. However, permission to use this material for any other purposes must be obtained from the IEEE by sending a request to pubs-permissions@ieee.org.}
}

\markboth{ IEEE INTERNET OF THINGS JOURNAL}{Roberg \MakeLowercase{\textit{et al.}}: DPNet: Dynamic Pooling Network for Accurate and Efficient Size-Aware Tiny Object Detection}

\maketitle

\begin{abstract}
In unmanned aerial systems, especially in complex environments, accurately detecting tiny objects is crucial. Resizing images is a common strategy to improve detection accuracy, particularly for small objects. However, simply enlarging images significantly increases computational costs and the number of negative samples, severely degrading detection performance and limiting its applicability.
This paper proposes a \textit{Dynamic Pooling Network (DPNet)} for tiny object detection to mitigate these issues. \textit{DPNet} employs a flexible down-sampling strategy by introducing a factor (\textit{df}) to relax the fixed down-sampling process of the feature map to an adjustable one. 
\M{Furthermore, we design a lightweight predictor to predict \textit{df} for each input image, which will be used to decrease the resolution of feature map in backbone.} 
Thus, we achieve input-aware down-sampling. 
We design an \textit{Adaptive Normalization Module (ANM)} to make a unified detector well compatible with different \textit{dfs}. At the same time, we also design a guidance loss to supervise the predictor's training.
DPNet realizes the dynamic allocation of computing resources to trade off detection accuracy and efficiency through this. Experiments on the TinyCOCO and TinyPerson datasets show that our DPNet can save over 35\% and 25\% GFLOPs, respectively, while maintaining comparable detection performance.The code will be made publicly available.
\end{abstract}

%
\IEEEpeerreviewmaketitle


\section{Introduction}

\begin{figure}[!t]
\centering
\includegraphics[width=.48\textwidth]{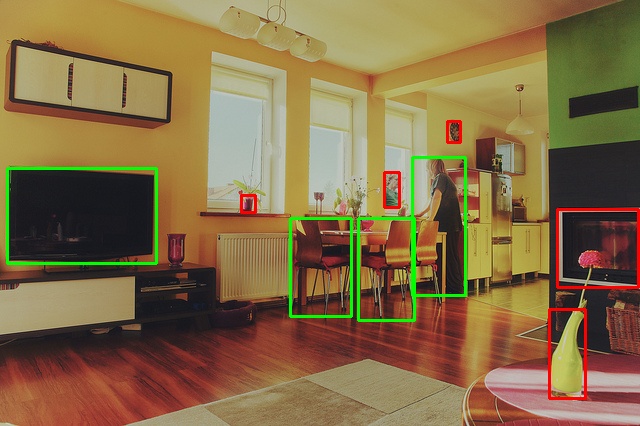}
\caption{Small objects can be better detected through resizing up the images. Objects in green boxes are correctly detected by both the detector trained on the original images, and the detector trained on enlarged images. Objects in red boxes are only correctly detected by the detector trained with enlarged images.
}
\label{fig:comp_139_scale8}
\vspace{-10pt}
\end{figure}

\begin{figure*}[!t]
\centering
\includegraphics[width=\textwidth]{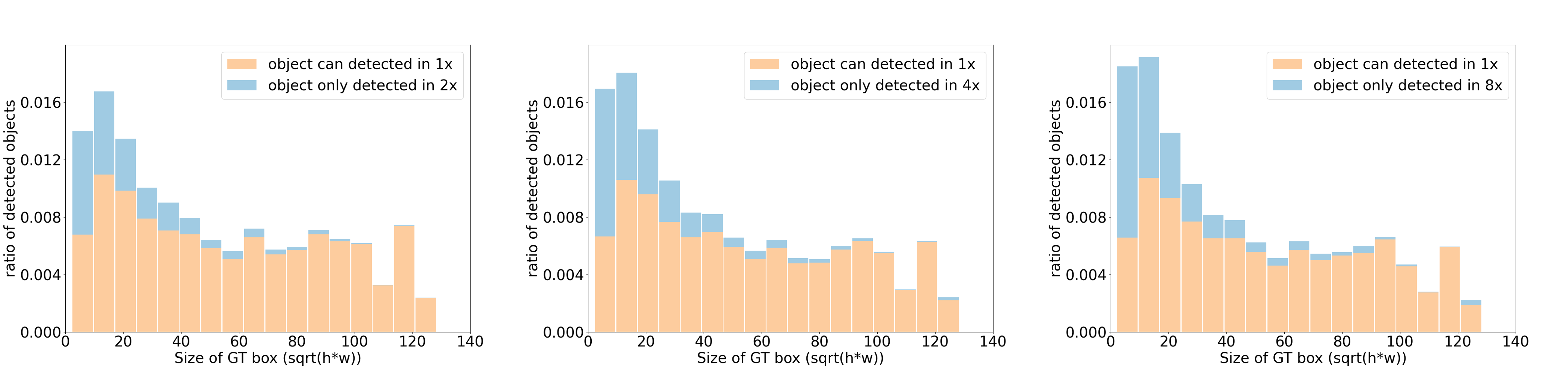}
\caption{
\M{
The performance gain over different objects' scales by enlarging (resizing up). We train and inference on origin images (1x) and enlarged images (2x, 4x, 8x) respectively. Then we statistic the objects that can be detected by both 1x input and enlarged input (orange), and the objects that can only be detected by enlarging the input (blue).
This figure indicates two points:
1) The performance gain from enlarged images mainly comes from tiny/small objects.
2) For medium and large objects, blindly enlarging input images does not improve performance, i.e. there is computational redundancy in enlarging. However, for tiny objects, the higher the enlarging level, the greater the performance gain.
Therefore, we need a strategy that can dynamically adjust the resolution for different input.
}
}
\label{fig:res_comp_hist}
\end{figure*}

\IEEEPARstart{T}{iny} object detection \M{(TOD)~\cite{DBLP:conf/wacv/YuGJYH20,ozge2019power,DBLP:conf/icassp/JiangYPGH21,bosquet2021stdnet,DBLP:conf/wacv/GongY0PZH21,yu20201st,cheng2022towards, liang2019small, duan2019detecting, fang2023enhancing}} has long been a challenging research direction in object detection. TOD aims to accurately detect objects with small sizes and a low signal-to-noise ratio.
Due to a large number of tiny objects in the real scene, tiny object detection has a wide range of application prospects. It plays an indispensable role in various fields such as \M{automatic driving~\cite{zheng2023visual,yang2018real}, smart medical treatment~\cite{rezaei2014fuzzy}, defect detection~\cite{wu2022digital}, and aerial image analysis\M{~\cite{segl2001detection,wang2020detection,liu2020uav,chen2023high,cheng2022tiny}}},particularly in drone system image recognition and analysis. Deep learning has injected fresh blood into tiny object detection research in recent years. With the rapid development of deep convolutional neural networks (CNNs), research in \M{object detectors~\cite{DBLP:conf/cvpr/LinDGHHB17, DBLP:focalloss, DBLP:conf/iccv/TianSCH19,DBLP:sparsercnn}} has made significant strides.
For extremely small sizes compared with objects in MS-COCO~\cite{DBLP:COCO}, one common approach is to resize pre-training objects to tiny object scales, as demonstrated in methods like Scale Match (SM) ~\cite{DBLP:conf/wacv/YuGJYH20} and its enhanced version SM+ ~\cite{DBLP:conf/icassp/JiangYPGH21}. However, these methods suffer information loss when downsizing images. Another common way is to enlarge the images' size to make \M{tiny objects larger~\cite{yu20201st,ozge2019power}}. Taking Fig.~\ref{fig:comp_139_scale8} as an example, when we downsizing the input image, the detector will be able to detect small objects better. Therefore, appropriately resizing up the input image or reducing the down-sampling operation can improve the detector's performance on tiny objects. However, resizing up the images also introduces some disadvantages: 1) increasing computational costs; 2) more negative samples; and 3) redundancy, as shown in Fig.~\ref{fig:res_comp_hist} (for some objects, it is unnecessary to resize the image to a larger size).

In order to obtain a trade-off between speed and performance, and eliminate the redundancy caused by enlarging all images, we design a dynamic neural network with variable $df$s (denoted by $df$), called Dynamic Pooling Network (DPNet). 
Different from the traditional (static) neural network~\cite{he2016deep,huang2017densely,simonyan2014very,ren2015faster} that uses the same parameters of down-sampling for all input images for inference during testing, our DPNet applies adaptive inference in the term of $df$. 
In order to get an appropriate $df$ of the input, we introduce a down-sampling factor predictor (DFP), which is embedded into the backbone of the detector. In practical terms, several different $df$s are set as candidates. The predictor is fed with an input feature map, produces a probability distribution over candidate $df$s as the output, and chooses the best $df$. 
Then in the next stage of the backbone, the feature map will be scaled by the selected $df$. DFP is lightweight enough to be ignored to calculate the computation cost.
For the detection task, the intrinsic difficulty level of the sample for recognition and the size and number of instances in the image may affect the amount of information required for correct detection. Therefore, when designing the DFP, we also introduce a branch to predict the statistical value of instances in the input image, which can help improve the rationality and effectiveness of DFP. By exploiting the proposed DPNet inference approach, we can excavate the spatial redundancy of each feature map from its size.

 
 In summary, our core contributions are as follows:

\begin{itemize}
\item We propose Dynamic Pooling Network (DPNet), which to our best knowledge, is the first attempt to introduce the dynamic neural network to object detection tasks to achieve the trade-off of detection performance and computation.
\item We design adaptive normalization module (ANM)
to solve the aggravated scale variation issue brought by the mixed scale training (MST) scheme. A down-sampling factor predictor (DFP) is adopted to choose $df$ adaptively during inference,
and we design a guidance loss to supervise its training.
\item Our DPNet can save over 35\% computing cost while maintaining the comparable detection performance on the TinyCOCO dataset. Experiments on different backbone networks validate the effectiveness of our method.
\end{itemize}

\section{Related Work}
Tiny object detection is a widely concerned problem in object detection, which attracts increasing research. The object scale influences the detection performance. Thus many works focus on the scale problems in object detection.
Dynamic neural networks can adaptively adjust models or input, which can help achieve the trade-off of computation and performance. In addition, to decrease computation, making the model lightweight is a traditional thought. Our method is a new paradigm that can work with the lightweight model method.
\subsection{Tiny Object Detection}
Extensive research has been carried out on tiny object detection from various aspects. Low resolution, weak signals, and high noise increase the difficulty of tiny object detection, limiting its research and application. To get reliable tiny object feature representation, Yu \textit{et al.}~\cite{DBLP:conf/wacv/YuGJYH20} and Jiang \textit{et al.}~\cite{DBLP:conf/icassp/JiangYPGH21} use methods that align the scale distribution of the pre-training dataset and the target dataset. Gong \textit{et al.}~\cite{DBLP:conf/wacv/GongY0PZH21} designed effective fusion factors of adjacent layers in FPN~\cite{DBLP:conf/cvpr/LinDGHHB17}. Several methods recover information on low-resolution objects by way of Super-Resolution (SR). Employing large-scale SR features, EFPN~\cite{DBLP:journals/corr/abs-2003-07021} extends the original FPN with a high-resolution level specialized for small-sized object detection. To get super-resolution features, Noh \textit{et al.}~\cite{DBLP:conf/iccv/NohBLSK19} uses high-resolution object features as supervision signals that match the relevant receptive fields of input and object features.
~\cite{DBLP:conf/eccv/RFLA} proposes a Gaussian Receptive Field based Label Assignment (RFLA) strategy for tiny object detection.

\subsection{Scale Based Detection}
The object scale has an important influence on the accuracy of the detection task. And it must also be considered in the design process of our method. Therefore, the object detection method that focuses on the scale problem needs to be discussed in the following.
 Large-scale variation across object instances is a challenging problem in object detection. Handling this problem can improve the ability of the detector to deal with the object detection task on various scales, especially for objects of extreme size, 
 \eg tiny objects.
 A common strategy to improve the detection methods is the multi-scale image pyramid~\cite{DBLP:conf/cvpr/HuangRSZKFFWSG017, DBLP:conf/cvpr/LiuQQSJ18}. Based on this scheme, SNIP~\cite{DBLP:conf/cvpr/SinghD18} and SNIPER~\cite{DBLP:conf/nips/SinghND18} apply a scale regularization method to multi-scale training, which guarantees the sizes of objects fall into a fixed scale range for various resolution images. Another normalization method, TridentNet~\cite{DBLP:conf/iccv/LiCWZ19}, constructs parallel multi branches with different receptive fields to generate scale-specific feature maps. These feature maps have a uniform representational power. 
 Instead of using multiple resolution input images and receptive field convolution kernels, SSD~\cite{DBLP:conf/eccv/LiuAESRFB16} and MS-CNN~\cite{DBLP:conf/eccv/CaiFFV16} utilize multiple spatial resolution feature maps to detect objects with different scales. They assign small targets to the bottom layers with high resolution, and large targets to the up layers with low resolution. To enhance the semantic representation of low-level features at the bottom layers, TDM~\cite{DBLP:journals/corr/ShrivastavaSMG16} and FPN~\cite{DBLP:conf/cvpr/LinDGHHB17} further introduce a top-down pathway and lateral connections that merge deep layers and shallow layers. On the basis of FPN, PANet~\cite{DBLP:conf/cvpr/LiuQQSJ18} adds a bottom-up path and proposes adaptive feature pooling to enhance the representational power of feature layers.

\subsection{Dynamic Neural Networks}
As an emerging topic in deep learning, dynamic neural networks have the advantages of high efficiency, great adaptability and strong representative ability, etc.
To meet varying computing resource demands, \cite{DBLP:conf/iclr/HuangCLWMW18, DBLP:conf/icpr/Teerapittayanon16, DBLP:conf/icml/BolukbasiWDS17,DBLP:journals/pami/HanHSYWW22} perform inference with dynamic network depths, which allows "easy" samples to be output at shallow exit without exceuting deeper layers. \cite{DBLP:conf/iclr/DehghaniGVUK19, DBLP:conf/iclr/ElbayadGGA20} implement halting scheme on Transformers~\cite{DBLP:conf/nips/VaswaniSPUJGKP17} to achieve dynamic network depth on NLP tasks. Yu \textit{et al.}~\cite{DBLP:conf/iclr/YuYXYH19, DBLP:conf/iccv/YuH19} present switchable batch normalization and in-place distillation to train a neural network executable at different widths. According to on-device benchmarks and computing resource constraints, the trained network, named slimmable neural network, can dynamically adjust its width. DS-Net~\cite{DBLP:conf/cvpr/LiWWLLC21} is also a network with variable width, which achieves dynamic routing for different samples by learning a slimmable supernet and a dynamic gating mechanism. RANet~\cite{DBLP:conf/cvpr/YangHCSDH20} implements resolution adaptive learning in deep CNNs for performing classification tasks efficiently. DRNet~\cite{DBLP:DRNet} dynamically adjust the resolution of input images for classification to achieve the trade-off of efficiency and classification accuracy. Our method first adopts these dynamic neural network ideas into detection tasks for wider application.

\subsection{Light Weight Models}
The efficient lightweight CNN models, such as MobileNet~\cite{DBLP:mobilenet,DBLP:mobilenetv2} and ResNeXt~\cite{DBLP:Resnext} are widely utilized in the object classification task, which can be used as the backbone of object detectors. As we know, the backbone for feature extraction occupies most computation in detectors, so a lightweight backbone is the main research hotspot in the efficient network.
MobileNet-v1~\cite{DBLP:mobilenet} adopts depthwise separable convolution to reduce parameter and computation without losing detection performance. MobileNet-v2~\cite{DBLP:mobilenetv2} introduces inverted residual block and linear bottleneck to improve the performance further. ResNeXt~\cite{DBLP:Resnext} introduces aggregated transformations, which replaces the residual structure of ResNet~\cite{DBLP:resnet} by stacking blocks with the same topology in parallel. These models save computation by simplifying the network, however, our methods reduce that by dynamic controlling the size of the input feature map, which can work together with these lightweight models.

\begin{figure}[t]
\centering
\includegraphics[width=.48\textwidth]{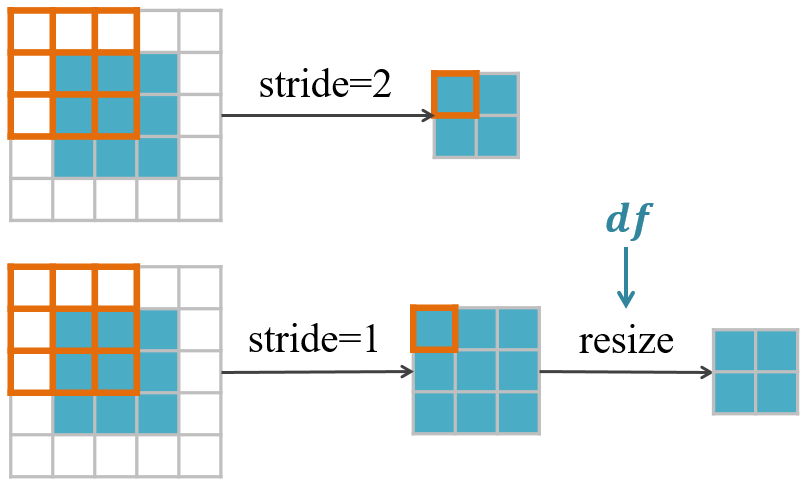}
\caption{The implement of the dynamic down-sampling. The down-sampling of feature maps is often implemented by setting the convolution kernel's strides, thus making the resolution of feature maps in the backbone fixed (top of the figure). We replace this process by resizing the features with the factor $df$ through \M{bilinear interpolation (bottom of the figure). while \textit{df}$\textless$0.5, the required GFLOPs of subsequent network modules will reduce}.}
\label{fig:stride2ds}
\end{figure}

\section{Dynamic Pooling Network}

In this section, we first present the overall architecture of our approach. Next, we elaborate on the training procedure, the Adaptive Normalization Module (ANM) and the down-sampling factor predictor (DFP) individually. 



Enlarging the image can significantly improve the overall performance of tiny object detection as showing  in Table~\ref{tab:different depth, width and size} (a) \M{(in Section \ref{subsec: analysis})} and describing in ~\cite{ozge2019power,yu20201st}, but at the same time it will bring a large computational cost. 
Therefore in this paper, We choose the enlarged images as the detector's input to take advantage of the information gained from enlarging. And then an adaptive down-sampling process is carried out on the feature map of the detector's backbone, reducing redundancy computation while retaining necessary information. 
As shown in Fig.~\ref{fig:stride2ds}, the down-sampling factor of the traditional static network structure is fixed(\textit{df}=0.5), which is implemented by a convolution or pooling layer with strides=2. To enable down-sampling at any \textit{df}, We perform bilinear interpolation on the feature map with the given down-sampling factor \textit{df}. 
\MM{Since the computational complexity of subsequent network modules is highly related to the resolution of the input feature map}, using different \textit{df} can control the computation cost in the network. 
In addition, the difficulty and the required minimum resolution of detecting objects in different images are different. For some images (e.g. size of objects inside is tiny), good results can only be obtained on high-resolution feature map, so we use a larger \textit{df} (\eg 0.5). Conversely, for some images (e.g. size of objects inside is large), good performance can also be achieved on lower-resolution feature map, which means there is some computational redundancy, so we use a smaller \textit{df} (e.g. 0.25). 
\begin{figure*}[ht]
\centering
\includegraphics[width=\textwidth]{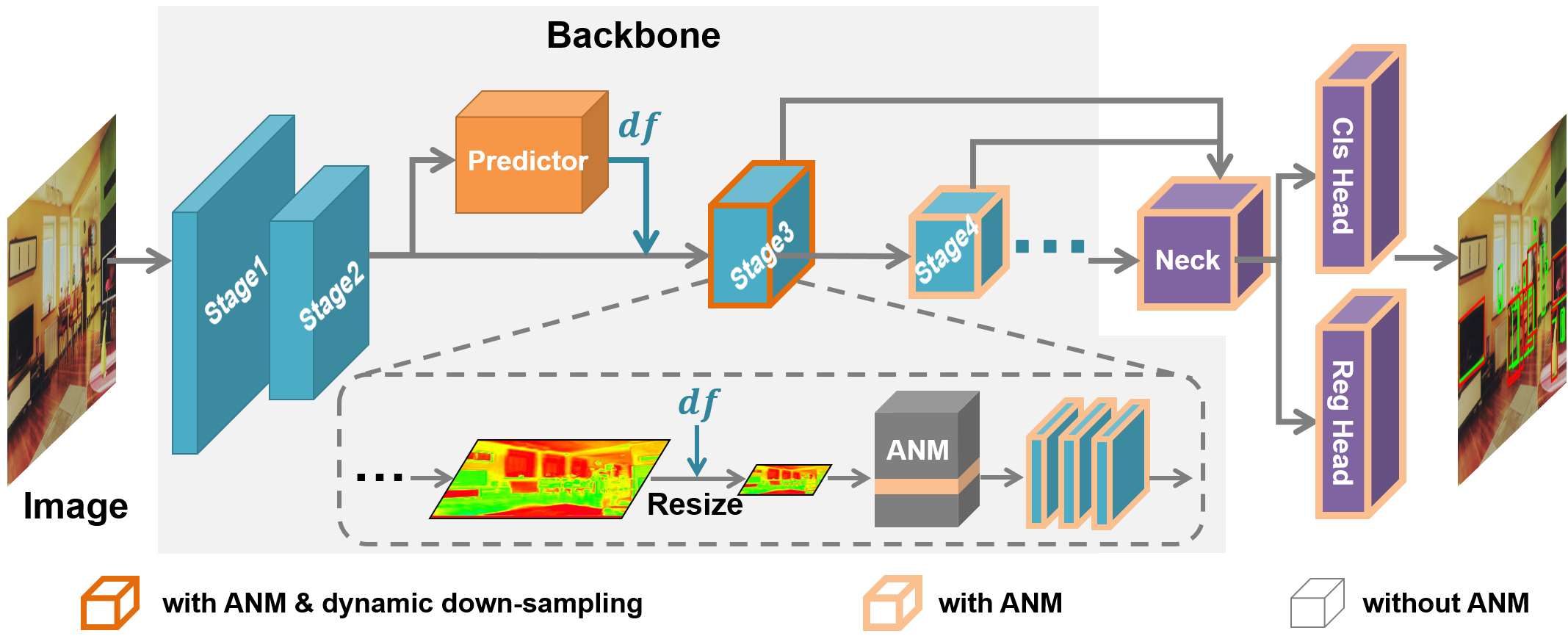}
\caption{Framework of DPNet. DFP is inserted into the backbone as a plug-in and guides a stage's the $df$ selection. After the feature is rescaled by $df$, all normalization layers in the detector are replaced with ANM. The ANM will switch to the corresponding normalization layer for the feature map. The blue cubes are the stages of the backbone, the orange one is the proposed DFP, and the purple ones are the neck and head parts of the detector. (Best viewed in color)}
\label{fig:framework}
\vspace{-10pt}
\end{figure*}
Therefore, by dynamically selecting the down-sampling factor  \textit{df} based on the input image, we can achieve a trade-off between performance and computational cost. 

The whole work can be divided into two tasks: 
1) Train a detector that can well handle multiple \textit{df}s, which means it can achieve high detection performance for all different \textit{df}s. To this end, we adopt a mixed down-sampling factor training policy (detail in Section~\ref{sec: DFP training}) and design an Adaptive Normalization Module (ANM, detail in ~\ref{sec: ANM}) to make a unified detector compatible with different \textit{df}s. 
2) Choose the smallest \textit{df} that still allows the detector to achieve strong detection performance on given input image under limited computational resources. For this purpose, We build a lightweight predictor (detain in ~\ref{sec: DFP}) embedded into detector to adaptively select the appropriate \textit{df} according to the input image to decrease the feature map resolution.


 
In this section, we will introduce the training procedure. As shown in Fig.~\ref{fig:framework}, our DPNet mainly consists of two components, a CNNs-based detector (such as RepPoints~\cite{DBLP:reppoints}) and a plugin-in DFP.  To achieve a better performance, We optimized the basic detector network and $df$ predictor one by one for better performance, as illustrated in Algorithm~\ref{al:1}.

\noindent\textbf{Mixed down-Sampling factor Training (MST).}

Note that there can be an arbitrary value for $df$ candidates from $0$ to $1$, making it difficult, also meaningless,
As a simplification strategy under practical requirement, we choose $m$ discrete $df$s $\{d_1, d_2,...,d_m\}$ to shrink the exploration range, where $d_m$ means the default down-sampling in a common detector. 
\M{
To achieve high detection performance over all different \textit{df}s, one naive solution is training a separate network for each \textit{df}, without any weight sharing between networks (the line of $SF$ in Table~\ref{tab:ANM}). However, maintaining multiple networks in this way is neither elegant nor practical for deployment. Therefore, a more common way is to train a single network under multiple \textit{df}s (mixed downsampling factor training) to enable it to handle various \textit{df}s (the line of $MF$ in Table~\ref{tab:ANM}).}

\subsection{DPNet training} \label{sec: DFP training}

\begin{algorithm}[H]
\caption{Training of DPNet \textit{M}}\label{al:1} 
\begin{algorithmic}[1]
	\REQUIRE Training set $D_{train}$, DPNet \textit{M}
    \STATE // {Train the detector \textit{D}}
    \STATE Initialize shared convolutions and fully-connected layers of \textit{D}.
    \STATE Initialize independent normalization layer of each $d_j$ in $df\_list$: $\left[d_1, d_2,...,d_m\right]$
    \FOR{$i = 1, \dots, n_{iters\_pretrain}$} 
       \STATE Image $x$ and label $y$ of a mini-batch.
       \FOR{$d_j$ in $df\_list$}
       \STATE Switch to the normalization parameters of $d_j$.
       \STATE Resize the feature map of $x$ with $d_j$.
       \STATE Compute
       \MM{$\mathcal{L}_{MST_j}
       = {L}_{cls_j}+\alpha{L}_{reg_j}$ with $d_j$, Eq.~\ref{Eq:MST_cls_reg}}.
    \ENDFOR
    \STATE Compute loss over all \textit{df} $\mathcal{L}_{MST}= \sum\limits_{j} \mathcal{L}_{MST_j}$\MM{, Eq.~\ref{Eq:MST_sum_j}}. \\
    \STATE Backward with $\mathcal{L}_{MST}$ and update parameters of detector $D$.
    \ENDFOR
    \STATE // Train the DFP predictor \textit{P}
    \STATE Initialize \textit{P}
    \STATE Freeze the parameters of \textit{D}
    \FOR{$i = 1, \dots, n_{iters\_pretrain}$}
    \STATE Compute
    $\mathcal{L}_{P}$\MM{, Eq.~\ref{Eq:predictor_loss}}, backward with it and update parameters of $P$.
    \ENDFOR 
   \RETURN \textit{M} 
\end{algorithmic} 
\end{algorithm}


As Fig.~\ref{fig:mst} showing, when training the DPNet detector, the same image is input to the network with m different $df$s $\{d_1, d_2,...,d_m\}$ conducted forward m times, where the loss function is calculated as follows:
\begin{figure*}[t]
\centering
\includegraphics[width=0.9\textwidth]{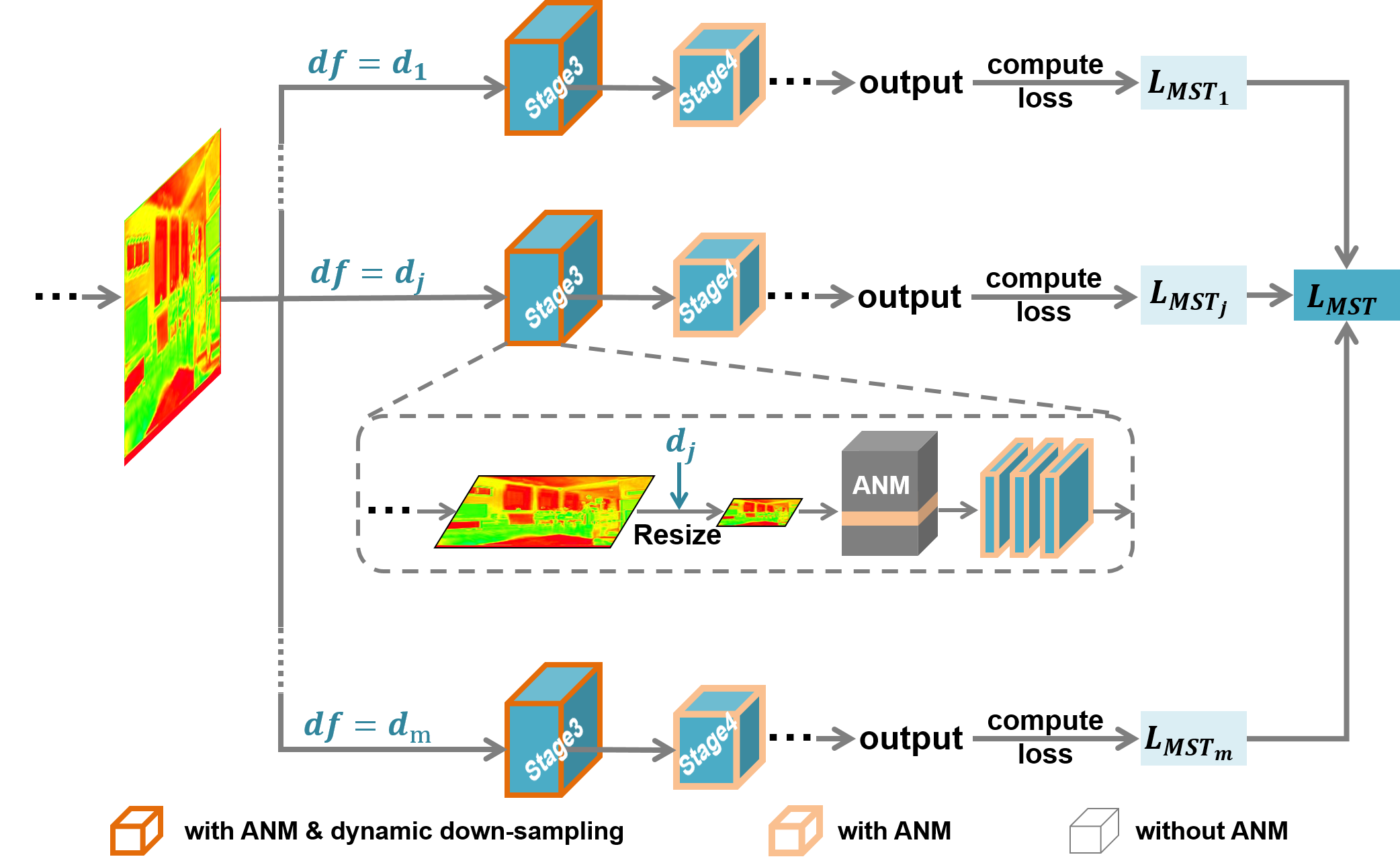}
\caption{ The pipeline of Mixed Scale Training. To achieve high detection performance over all
different $df$s, the same image is input to the network with m different $df$s $\{d_1, d_2,...,d_m\}$ conducted forward m times parallel and the loss are summed as Eq.~\ref{Eq:MST_sum_j}}
\label{fig:mst}
\end{figure*}

\begin{equation}
\begin{aligned}
\MM{\mathcal{L}_{MST_j}
= {L}_{cls_j}+\alpha{L}_{reg_j},}
\label{Eq:MST_cls_reg}
\end{aligned}
\end{equation}
where $\mathcal{L}_{MST_j}$ corresponds to the loss function of the down-sampling factor $d_j$, weighted and summed by the classification loss and the regression loss.
Each iteration will be updated by the loss summation of the reverse conduction computed by the network under m $df$s for the forward conduction:
\begin{equation}
\begin{aligned}
\mathcal{L}_{MST}= \sum\limits_{j=1}^{m} \mathcal{L}_{MST_j},
\label{Eq:MST_sum_j}
\end{aligned}
\end{equation}

\M{However, simply using MST policy causes significantly worse performance in most \textit{df} (Table~\ref{tab:ANM}) relative to the $SF$ policy that adopt a single \textit{df}  in both training and inference. This indicates that the network has not fully grasped the ability to handle multiple \textit{df}s for inference.  We conjecture that this is because feature maps produced by different \textit{df}s differ substantially. In a perspective, the feature maps produced by different \textit{df}s can be regarded as belonging to different domains. Therefore, inspired by~\cite{chang2019domain,li2018adaptive}, we design an Adaptive Normalization Module (ANM) to handle the domain issue, making a unified detector compatible with different \textit{df}s.}

\noindent\textbf{DFP Training.}
During the DFP training, the parameters of the detector network are fixed. 
\M{The DFP aims to automatically select the smallest \textit{df} that can achieve the best performance under limited computational resources for the input image. Since there are $m$ choices for \textit{df}, we model this problem as an $m$-class classification problem.} {The DFP takes the feature map  $M$ as input and passes through a lightweight structure to predict the selecting probability $p_s \in \mathbb{R}^m$ over $m$ \textit{df}s:

\begin{equation}
\begin{aligned}
DFP(M)=p_s
= [p_{d_1}, p_{d_2}, \dots,p_{d_m}],
\label{Eq:predictor}
\end{aligned}
\end{equation}
To obtain $df$ predicted by DFP, the factor $d_j$ corresponding to the highest probability $p_{d_j}$ is selected as the $df$ fed to the next stage of the backbone:

\begin{equation}
\begin{aligned}
\M{df = argmax_{d_j}\ p_{d_j}}
\label{Eq:d_j select}
\end{aligned}
\end{equation}

Here is an extremely important issue for training DFP: how to obtain supervision for training DFP classifier. This issue will be addressed in Section~\ref{sec: DFP}. Here we assume that we already have the supervision $y_{d_j}$.
Then a traditional multi-label classification loss $\mathcal{L}_{df}$ is defined as follows:
\begin{equation}
\mathcal{L}_{df}=\sum_{j=1}^{m} -y_{d_j} \log \left(p_{d_j}\right)-\left(1-y_{d_j}\right) \log \left(1-p_{d_j}\right).
\label{Eq:$df$_loss}
\end{equation}

Since the prediction of \textit{df} is closely related to the objects' size in the image, we added a statistical branch to assist the training of DFP. This statistical branch attempts to predict the statistical information $\hat{v}$ (e.g. mean, min, max) about the object's size in the image. This allows DFP to learn the scale information of the objects in the image, so as to better predict \textit{df}.
\MM{The Smooth L1 loss~\cite{girshick2015fast} is performed between predicted $\hat{v}$ and target label $v$ as ($\beta_{sta}$ is set as $1/9$, following~\cite{girshick2015fast})}:}

\MM{
\begin{equation}
\begin{aligned}
\mathcal{L}_{sta}&=SmoothL1(\hat{v}, v) \\
                 &=\left\{
 \begin{aligned}
0.5(\hat{v} - v)^2/\beta_{sta}, &\ if\ |\hat{v}- v| \leq \beta_{sta}, \\
|\hat{v} - v| - 0.5*\beta_{sta}, &\ otherwise. 
\end{aligned}
\right.
\label{Eq:predictor_l1}
\end{aligned}
\end{equation}
}

The objective function of DFP is a weighted summation of two losses:

\begin{equation}
\begin{aligned}
 \MM{\mathcal{L}_{P}=\mathcal{L}_{sta} + \lambda \mathcal{L}_{df},}
\label{Eq:predictor_loss}
\end{aligned}
\end{equation}
where \MM{$\lambda=1.0$ (by default in this paper)}.


\begin{figure}[t]
\centering
\includegraphics[width=.48\textwidth]{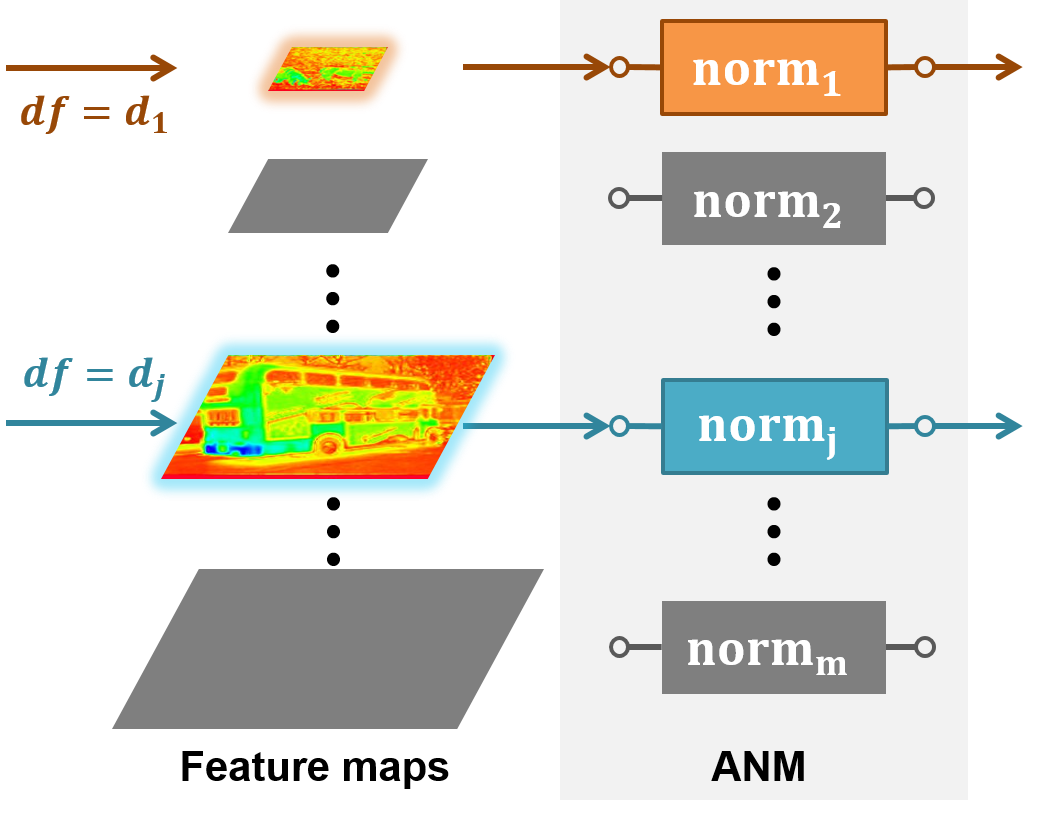}
\caption{The mechanism of the ANM. Different features scaled by different $df$s will be fed into ANM, and ANM  will switch to the corresponding normalization layer for the $df$.}
\label{fig:ANM}
\end{figure}

\subsection{Adaptive Normalization Module} \label{sec: ANM}
\M{Considering the limitation of computational resources, we train a shared detector with the MST strategy. It can inference under different \textit{df}s, enabling it to adapt to different computational loads for inference on various devices. 
To handle the distributions' differences in feature map introduced by different \textit{df}s, inspired by~\cite{chang2019domain,li2018adaptive}, we set different normalization layers for different \textit{df}s and auto switch these layers with given \textit{df}, which named Adaptive Normalization Module (ANM).
}

Normalization plays an important role in solving the internal covariate shift problem in deep neural network learning ~\cite{DBLP:BatchNormalization}.
\MM{It can encode} condition information to feature representations ~\cite{DBLP:FiLM,DBLP:conf/iclr/LiWS0H17}.
It also can accelerate convergence and improve stability ~\cite{DBLP:journals/corr/RadfordMC15} through normalizing the input feature by re-centering and re-scaling:

\begin{align}
  y^{\prime}=\gamma \frac{y-\mu}{\sqrt{\sigma^{2}+\epsilon}}+\beta,
  \label{eq: norm}
\end{align}
where $y$ is the input to be normalized and $y^{\prime}$ is the output. $\gamma$, $\beta$ are learnable scale and bias, $\mu$, $\sigma^{2}$are the mean and variance of input.

To train a network with variable $df$s, \textit{ANM} privatizes all normalization layers for each switch in a union network. It solves the problem of feature aggregation inconsistency between different switches by independently normalizing the feature mean and variance during testing. The scale and bias in \textit{ANM} may be able to encode conditional information of $df$ configuration of the current switch.
Moreover, in contrast to incremental training, with \textit{ANM} we can jointly train all switches at different $df$s; therefore all weights are jointly updated to achieve better performance.

As shown in Fig.~\ref{fig:ANM}, \textit{ANM} decouples the normalization for each $df$, \M{$d_j\in\{d_1, d_2,...,d_m\}$} and chooses the corresponding normalization layer to normalize the features:

\begin{align}
 y^{\prime}=\gamma_{j} \frac{y-\mu_{j}}{\sqrt{\sigma_{j}^{2}+\epsilon}}+\beta_{j}, j \in\{1,2, \ldots, m\},
\end{align}
 where $\epsilon$ is a small number for numerical stability, $\mu_{j}$ and $\sigma_{j}^{2}$ are private averaged mean and variance from the activation statistics under separate $df$s; $\beta_{j}$ and $\gamma_{j}$ are private learnable scale weights and bias.

\subsection{Down-sampling Factor Predictor} \label{sec: DFP}

In order to choose the $df$s of a feature map, a $df$s predictor (DFP) network is designed. \M{ The goal of DFP is to predict an appropriate $df$ that reduces the calculation consumption as much as possible while maintains a high detection performance, as described in Section~\ref{sec: DFP training}.
In this section, we will introduce how to obtain supervision $y_{d_j}$ and $v$ for training DFP.}

\begin{algorithm}[H]
\caption{Guidance loss of DFP \textit{P}}\label{al:2} 
\begin{algorithmic}[1]
	\REQUIRE Training set $D_{train}$, DPNet \textit{M}
    \FOR{$I_i$ in $D_{train}$}
       \FOR{$d_j$ in $df list$}
       \STATE Get the predicted result box $bboxes_i$ for $D(I_i)$.
       \STATE Get $pos_i$, $neg_i$ samples by assigning $bboxes_i$.
       \STATE $neg_i=sorted(NMS(neg_i))$.
        \STATE $pos=[]$
        \FOR{$gt$ in $I_i$}
        \STATE \M{$pos_*=argmax_{pos_* \in pos_i} IoU(pos_i, gt)$}
        \STATE $pos=pos \cup pos_*$
        \ENDFOR
        \STATE $k=length(pos)$
        \STATE $neg=top_k(neg_i)$
        \STATE 
       $\mathcal{L}_{i}^{(j)}=$ loss of $pos$ and $neg$.
    \ENDFOR
    \ENDFOR
   \STATE\RETURN $\mathcal{L}_{i}^{(j)}$ 
\end{algorithmic} 
\end{algorithm}

\M{\noindent\textbf{Guidance loss.} How to obtain the supervision $y_{d_j}$ for $df$ classification?} The category label of the DFP is related to the detector, that is to say, when the images are input to the detector with different $df$s, the $df$ with high detection performance should be \M{labeled as 1 (positive)}. Therefore, each image can be input to the detector to obtain the label of the DFP. If the performance is high, the $df$ is labeled as 1 for the predictor, and if the performance is too poor compared to other $df$s, it is labeled as 0 (negative). 
To measure the performance of a detector, one of the most useful metric is AP \MM{(average precision~\cite{zhu2004recall})}. If a $df$ gets the best AP performance among all $df$s, then that $df$ is the corresponding category label of this image in the DFP. However, it is impractical to use the AP metric to label each image due to AP of a single image is not representative. 
Another method is to calculate the loss function directly. The design of the loss function needs to be as relevant and close to the AP as possible. So much so that after entering the detector, we can directly determine the detection performance of the input image at a certain $df$ based on this loss value.
\M{Therefore, in this paper, we designs a loss called the \textbf{guidance loss} noted as $L^{(j)}$, which is more closely related to the AP compared to the original loss of the detector. 
The guidance loss is calculated in Algorithm~\ref{al:2}.}
The purpose of the guidance loss is not for optimizing the detector, but for obtaining supervised information about the DFP, which is simply designed as a function that is negatively correlated with the AP.
The label $y_{d_j}$ in Eq~\ref{Eq:$df$_loss} is defined in the following way:
\M{
\begin{equation}
 y_{d_j}=\left\{
 \begin{aligned}
1, &\ if\ \frac{{L}^{(j)}}{min({L}^{(j)})} \leq T, \\
0, &\ otherwise. 
\end{aligned}
\right. j = 1, 2, ... m
\label{Eq:$df$_loss_label}
\end{equation}
}
where $T$ is is set to 1.1 by default, ${L}^{(j)}$ is the loss value calculated when $df=d_j$ by feeding the input image into the detector trained in detector training step (see Algorithm~\ref{al:1}), $min({L}^{(j)})$ is the minimum of all ${L}^{(j)}$s ($j = 1, 2, ... m$).

\begin{figure}[t]
\centering
\includegraphics[width=.48\textwidth]{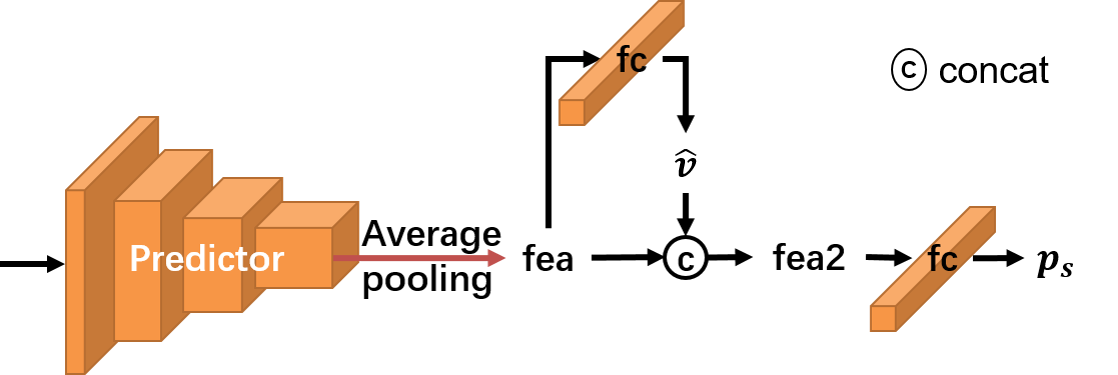}
\caption{The structure of DFP. DFP refers to and simplifies the structure of ResNet in its design. Additionally, a statistic branch is added to improve the effectiveness of prediction.}
\label{fig:predictor}
\end{figure}

\noindent\textbf{Statistic Branch.} The rescaled feature map should better adapt to the detector. To this end, a branch to predict the characteristics of instances in the input image is added to the DFP, called \textit{Statistic Branch}. Statistic Branch predicts a vector $\hat{v}$ containing the number of instances in the input image and the statistic values (\eg the mean, maximum and minimum values) of their objects' size. \textbf{These values of supervision $v$ can be easily obtained from the box annotation of the image.}
As shown in Fig.~\ref{fig:predictor}, the structure of the predictor consists of 4 basic blocks. Each basic block consists of two convolution layers with batch normalization and ReLU activations. 


\section{Experiments}
\subsection{Experimental Setting}
\noindent{\bfseries Dataset.} \textbf{TinyCOCO} dataset (MS-COCO 2017)~\cite{DBLP:COCO} is a widely-used benchmark to evaluate the detection performance of the tiny object detection tasks, which consists of 118k training images, 5k validation images and 20k testing images in 80 categories. All the studies are conducted on the validation set. TinyCOCO is the COCO100~\cite{DBLP:conf/wacv/YuGJYH20}, which resizes the shorter edge of each image to 100 and keeps the height-width ratio unchanged. The mean of objects' size in TinyCOCO is the same as the definition of the tiny object in~\cite{DBLP:conf/wacv/YuGJYH20}. \M{\textbf{TinyPerson} is a commonly used dataset for tiny object detection, specifically designed for drone aerial photography scenarios. The dataset is collected from high-quality drone videos and online images. It contains 72,651 annotated human objects with low visual resolution, and a total of 25,945 images cropped to a size of 640x512, aimed at validating the model's applicability in drone-based object recognition tasks. }

\noindent{\bfseries Evaluation Metric.} 
According to Tinybenchmark\cite{DBLP:conf/wacv/YuGJYH20}, we mainly use Average Precision(AP)~\cite{zhu2004recall} for the tiny-scale performance evaluation, which is the most popular metric in various object detection tasks. It reflects both the precision and the recall ratio of objects' detection results. The threshold value of IoU is set to $0.25$, $0.5$, $0.75$. Under the scenarios of the tiny-scale task, we pay far more attention to whether the target can be found than its location accuracy. Therefore, $IoU = 0.5$ naturally becomes the main evaluation criterion. Instead of the small, medium and large rules in the MS-COCO evaluation metric, we use tiny, small and reasonable as our basis for splitting scales. $\{mAP, AP_{50}, AP_{75}, AP_{s}, AP_{m}, AP_{l}\}$ is used as the reported evaluation metric. To evaluate the efficiency, we calculate the average GFLOPs. We use the average size of the validation set as the input size when calculating the FLOPs of the model under all different $df$s. Then, the proportion of each $df$ selected by the predictor is used as the weight to weighted sum each corresponding FLOPs. \M{For TinyPerson, we follow ~\cite{DBLP:conf/wacv/YuGJYH20} that divides tiny [2, 20] into 3 sub-intervals: tiny1[2, 8], tiny2[8, 12], tiny3[12, 20] and choose IoU= 0.5 as the threshold for evaluation.}

\begin{table}[t]
\centering
\begin{tabular}{ccccccc}
\hline
Train Method & Test $df$ &$mAP$  & $AP_{50}$  & $AP_{s}$ & $AP_{m}$ & $AP_{l}$ \\
\hline
\multirow{3}{*}{SF} & 0.5  & 29.4 & 47.7 & 22.5 & 44.2 & 53.6 \\
&0.33  &  25.3  & 41.5   & 17.4 & 42.9 & 52.9\\
&0.25  &  19.2  & 32.5  & 11.4 & 38.3 & 48.4 \\
\hline
\multirow{3}{*}{MF}& 0.5  & 24.1 & 40.4  & 17.9 & 37.1 & 45.9\\
&0.33 &  23.1 & 39.2  & 15.7 & 38.8 & 48.7\\
&0.25 &  21.3 & 36.2  & 13.6 & 37.4 & 50.3\\
\hline
\multirow{3}{*}{MF + ANM}&0.5  & 29.5 & 47.9  & 22.6 & 43.8 & 52.2 \\
&0.33 &  28.9  & 47.6   & 21.0 & 46.8 & 57.1\\
&0.25 &  27.4  & 45.0  & 18.5 & 47.2 & 59.5 \\
\hline
\end{tabular}
\caption{Influence of ANM. SF means training only with single $df=0.5$, MF means training detector with $df=0.5, 0.33, 0.25$.}
\label{tab:ANM}
\end{table}

\begin{table}[t]
\centering
\begin{tabular}{c|ccc}
\hline
Method & Flops & $mAP$  & $AP_{50}$ \\
\hline
Random-1& 107.80 GFLOPs & 28.90 & 47.50 \\
Random-2& 107.80 GFLOPs & 29.00 & 47.70 \\
Random-3& 107.80 GFLOPs & 28.90 & 47.60 \\
Random(mean)& 107.80 GFLOPs & 28.93 &  47.60 \\
DPNet(ours) & 107.80 GFLOPs & 29.70 & 48.50\\
\hline
\end{tabular}
\caption{DPNet vs. Random Factors}
\label{tab:Randomfactors}
\vspace{-10pt}
\end{table}

\begin{table*}[t]
\centering
\begin{tabular}{c|cccccccc}
\hline
Model &  Params & Flops& $\downarrow$ Flops & $mAP$  & $AP_{50}$ & $AP_{s}$ & $AP_{m}$ & $AP_{l}$ \\
\hline
\hline
\multicolumn{9}{c}{\textbf{ResNet-50}}\\
\hline
Faster R-CNN-ResNet-50 & 41.48 M & 56.72  & - & 18.7  & 34.2 & 10.7& 34.1 & 53.8\\
Faster R-CNN-ResNet-50 $^{\dagger}$ & 41.48 M & 183.96  & -  & 15.9 & 31.1 & 8.6 & 30.2 & 45.9 \\
Cascade R-CNN-ResNet-50 & 69.17 M & 84.36  & - & 18.0  & 32.3 & 9.7 & 34.4 & 53.7 \\
Cascade R-CNN-ResNet-50 $^{\dagger}$ & 69.17 M & 211.61   & -  & 30.7 & 46.5 & 23.0 & 47.8 & 54.1\\
RetinaNet-ResNet-50 & 37.74 M & 52.76  & - & 11.1  & 22.8 & 5.1 & 20.2 & 39.9 \\
RetinaNet-ResNet-50 $^{\dagger}$ & 37.74 M & 211.61  & -  & 26.3 & 41.7 & 19.7 & 41.2 & 47.8\\
\hline
RepPoints-ResNet-50 & 36.62 M & 41.99  & - & 18.7  & 34.2 & 10.7& 34.1 & 53.8\\
RepPoints-ResNet-50 $^{\dagger}$ & 36.62 M & 168.25  & -  & 29.4 & 47.7 & 22.5 & 44.2 & 53.6\\
\textbf{DPNet-ResNet-50 $^{\dagger}$(ours) }& 39.64 M & 107.80  & $\downarrow$35.93\% &29.7  & 48.5 & 22.0 & 46.5 & 56.6\\
\hline
\multicolumn{9}{c}{\textbf{ResNet-101}}\\
\hline
RepPoints-ResNet-101& 55.62 M & - & - & 18.4 & 32.7 & 9.6 & 36.5 & 52.3\\
RepPoints-ResNet-101$^{\dagger}$ & 55.62 M & 217.54  & - & 31.7 & 50.7 &24.6 & 47.3 & 56.1\\
\textbf{ DPNet-ResNet-101$^{\dagger}$(ours) }& 58.64 M & 157.78  & $\downarrow$27.47\%  & 31.6 & 50.8 & 23.8 & 48.8 & 59.8 \\
\hline
\multicolumn{9}{c}{\textbf{ResNeXt-50}}\\
\hline
RepPoints-ResNeXt-50  & 35.44 M & - & - & 19.5 & 34.5 &10.7&37.9&59.0\\
RepPoints-ResNeXt-50$^{\dagger}$ & 35.44 M & 168.25  & - & 29.9 & 48.3 & 23.0 &44.5&52.8\\
\textbf{ DPNet-ResNeXt-50$^{\dagger}$(ours)}& 38.46 M & 108.44  & $\downarrow$35.55\% & 29.8 &48.2 &22.9 &44.7& 53.1\\
\hline
\multicolumn{9}{c}{\textbf{MobileNet-v2}}\\
\hline
RepPoints-MobileNet-v2& 8.52 M & -  & -  &  14.9& 26.7 &8.3&27.3&43.8\\
RepPoints-MobileNet-v2$^{\dagger}$ & 8.52 M & 95.71  & -  & 25.2 & 41.7 &18.7&38.6&50.4\\
\textbf{ DPNet-MobileNet-v2$^{\dagger}$(ours)} &11.06 M& 56.08  &$\downarrow$41.40\% & 25.1& 41.9  & 17.5  & 40.0 & 52.3\\
\hline
\end{tabular}
\caption{Comparison with other model compression methods on TinyCOCO dataset.$^{\dagger}$ means that we enlarge the input images to 8 folds. We use RepPoints as our basic detector and perform on different backbones for comparison. AP is reported for detection performance and GFLOPs is reported for computation.}
\label{tab:big-table}
\end{table*}

\begin{table*}[t]
\centering
\begin{tabular}{c|ccccccc}
\hline
Model &  Params & Flops& $\downarrow$ Flops & $AP_{tiny}$ & $AP_{tiny1}$ & $AP_{tiny2}$ & $AP_{tiny3}$ \\
\hline
\hline
FreeAnchor~\cite{zhang2019freeanchor}  & 37.81 & 222.23 & - &  44.26 & 25.99 & 49.37 & 55.34 \\ 
Adap RetinaNet~\cite{DBLP:conf/wacv/GongY0PZH21} & 36.12 M & 177.52 & - &  46.56 & 27.08 & 52.63 & 57.88 \\ 
Grid RCNN~\cite{lu2019grid}	           & 64.32 & 189.15 & - &  47.14 & 30.65 & 52.21 & 57.21 \\ 
RetinaNet-S-$\alpha$~\cite{DBLP:conf/wacv/GongY0PZH21} & 36.12 M & 177.52 & - &  48.34 & 28.61 & 54.59 & 59.38 \\ 
RetinaNet-SM~\cite{DBLP:conf/wacv/YuGJYH20}  & 36.12 M & 177.52 & -&  48.48 & 29.01 & 54.28 & 59.95 \\ 
\hline
RepPoints    & 36.62 M & 158.92  & - & 44.51 & 27.94 & 51.22 & 54.85 \\
RepPoints-SM & 36.62 M & 158.92  & - & 48.88  & 33.79 & 55.10 & 58.48 \\
RepPoints-SM $^{\dagger}$ & 36.62 M & 635.66  & -  & 52.17 & 39.37 & 56.59 & 61.22 \\
\textbf{DPNet $^{\dagger}$(ours) }& 39.64 M &  471.74 & $\downarrow$25.79\% & 52.33 & 39.69 & 57.58 & 60.50 \\
\hline
\end{tabular}
\caption{Comparison with other methods on TinyPerson dataset.$^{\dagger}$ means that we enlarge the input images to 1.85 folds. All methods applies ResNet-50 as backone.}
\label{tab:big-table-tinyperson}
\end{table*}

\begin{table}[t]
\centering
\begin{tabular}{c|cccc}
\hline
Method & Backbone & Sched. & Size& $mAP$  \\
\hline
SSD& ResNet-50-FPN &1x & 1024 & 29.86 \\
R-FCN& ResNet-50-FPN &1x & 1024 & 52.58 \\
Faster-RCNN& ResNet-50-FPN &1x & 1024 & 60.46 \\
RepPoints& ResNet-50-FPN &1x & 1024 & 59.44 \\
\hline
\textbf{DAPNet(ours)}& ResNet-50-FPN &1x & 1024 & \textbf{63.55} \\
\hline
\end{tabular}
\caption{DPNet vs. other methods on DOTA dataset}
\vspace{-10pt}
\label{tab:dota}
\end{table}

\begin{table}[htbp]
\centering
\scriptsize
\begin{tabular}{l|cccc}
\hline
\textbf{Model} & \textbf{Input Size} & \textbf{$AP_{50:95}$} & \textbf{Params(M)} & \textbf{Flops (G)} \\
\hline
Faster-RCNN & 640x640 & 17.0 & 82.3  & 65.3 \\
YOLOV10m & 640x640 & 27.08 & 15.4  & 59.1 \\
YOLOV11m & 640x640 & 28.06 & 20.1  & 68.0 \\
YOLOV12m & 640x640 & 29.02 & 20.2  & 67.5 \\
\hline
\textbf{DPNet-MobileNet-v2} & 640×640 & \textbf{28.74} & \textbf{11.06} & \textbf{56.08} \\
\hline
\end{tabular}
\caption{Performance comparison on the VisDrone dataset (higher AP is better; lower Params/Flops are better)}
\label{tab:visdrone}
\vspace{-15pt}
\end{table}

\begin{table}[ht]
\centering
\begin{tabular}{c|ccccc}
\hline
Guidance loss &  $mAP$  & $AP_{50}$ & $AP_{s}$ & $AP_{m}$ & $AP_{l}$ \\
\hline
  & 29.8 & 47.8 & 21.6 & 48.2 & 60.4 \\
\Checkmark & 30.9 & 50.1 & 23.0 & 47.8 & 59.4\\
\hline
\end{tabular}
\caption{Performance upper bound influnce of guidance loss.}
\label{tab:guidance}
\vspace{-10pt}
\end{table}

\noindent{\bfseries Implement Details.} 
The implements of DPNet are based on MMDetection~\cite{mmdetection}. Same to the default setting of the object detection on MS-COCO. And the stochastic gradient descent (SGD~\cite{DBLP:series/lncs/Bottou12}) algorithm is used to optimize in 1$\times$ training schedule over 8 GPUs with a total of 16 images per minibatch (2 images per GPU). The learning rate is set to 0.02 and decays by 0.1 at the 8-th and 11-th epochs, respectively. For candidate \textit{$df$}s of the large detector, we choose \textit{$df$}s of [0.5, 0.33, 0.25] \M{(which means m=3)}.


\begin{table}[htb!]
\centering
\begin{minipage}[h]{\linewidth}
    \centering
    \par{(a) \\}
    \begin{tabular}{cccccc}
    \hline
    Size &  $mAP$  & $AP_{50}$ &  $AP_{s}$ & $AP_{m}$ & $AP_{l}$\\
    \hline
    (100, 167) & 18.7 & 34.2  & 10.7 & 34.1 & 53.8\\
    (150, 250)   & 23.4 & 40.2  & 14.6 & 41.4 & 57.7\\
    (200, 333)   & 24.8 & 43.1  & 16.5 & 41.7 & 54.8\\
    \hline
    \end{tabular}
\end{minipage}
\begin{minipage}{\linewidth}
    \centering
    \par{(b) \\}
    \begin{tabular}{cccccc}
    \hline
    Depth &  $mAP$  & $AP_{50}$ &  $AP_{s}$ & $AP_{m}$ & $AP_{l}$\\
    \hline
    34 & 17.1 & 30.6 & 08.9 & 32.5 & 53.7\\
    50 & 18.7 & 34.2  & 10.7 & 34.1 & 53.8\\
    101 & 18.4 & 32.7  & 09.6 & 36.5 & 52.3\\
    152 & 19.6 & 34.0 & 10.2 & 39.0 & 56.3\\
    \hline
    \end{tabular}
\end{minipage}
\begin{minipage}{\linewidth}
    \centering
    \par{(c) \\}
    \begin{tabular}{cccccc}
    \hline
    Width &  $mAP$  & $AP_{50}$  & $AP_{s}$ & $AP_{m}$ & $AP_{l}$\\
    \hline
    0.5 & 16.4 & 29.4  & 08.4 & 31.9 & 50.6\\
    1 & 18.7 & 34.2  & 10.7 & 34.1 & 53.8\\
    1.5 & 19.2 & 34.1  & 10.3 & 37.3 & 54.5\\
    \hline
    \end{tabular}
\end{minipage}
\caption{Performance comparison: RepPoints detector using Different input sizes or with backbone using different depths or sizes. \M{This shows that for tiny object detection tasks, increasing the resolution to leads to greater performance gains than increasing the depth or width of the network.}}
\label{tab:different depth, width and size}
\end{table}

\subsection{Ablation Study} 
\noindent{\bfseries Adaptive Normalization Module.} As shown in Table~\ref{tab:ANM}, we first train under the single $0.5$ $df$ and evaluate it on $\{0.5, 0.33, 0.25\}$ $df$s. The performance is 29.4, 25.3 and 19.2 mAP, respectively. With mixed factors, we down-sample the image with 0.5, 0.33 and 0.25 respectively and optimize the shared detector head with respective losses. However, we find the performance decreases a lot (over 5 points) which shows that same normalization layer is not suitable for different $df$s. So we use an adaptive normalization module, in which different $df$s have different normalizations. The performance increases to 29.5, 28.9 and 27.4 mAP, which makes the detector obtain good performance under all down-samplings.


\noindent{\bfseries Down-Sampling Factor Predictor.} In order to verify the effectiveness of our $df$s predictor, we use a random strategy for comparison (s shown in Table~\ref{tab:Randomfactors}). We random sample the $df$ based on the same GFLOPs to evaluate the performance. The operation is conducted three times and the mean performance is 28.93 mAP $\pm$ 0.05 and 47.60 AP$_{50}$ $\pm$ 0.05. The performance with our designed predictor is 1 point higher than those, which shows that the result does not come from randomness and our predictor really works.

\noindent{\bfseries Guidance Loss.} In order to verify the effectiveness of the guidance loss, the experiment first uses the optimization with and without the guidance loss as a $df$ directly to the detector, whose performance indicates the upper limit of the performance under the supervision. The higher the upper bound, the better the supervised information.
Table~\ref{tab:guidance} shows that the upper bound performance mmAP with the guidance loss is higher than that without the guidance loss by 1.1, which shows that the guidance loss is correlated with the detection performance, i.e., the lower the supervision loss, the higher the detection performance of the picture in the detector.

\begin{figure*}[t!]
\centering
\includegraphics[width=\textwidth, height=0.9\textwidth \textbf{}]{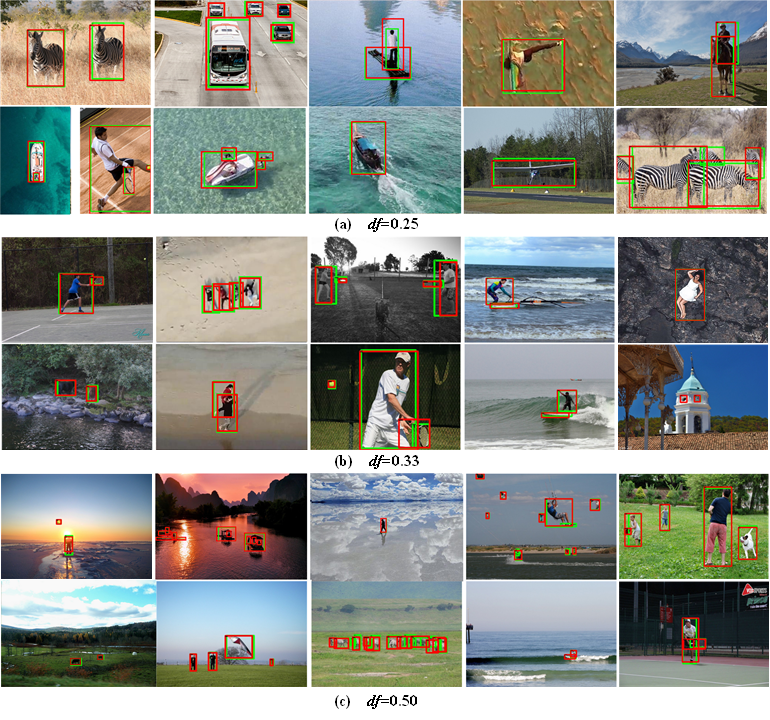}
\caption{Image visualization of DPNet on validation. Green boxes denote the ground truth and red boxes denote the predicted boxes. We conduct experiments on TinyCOCO, however visualize the results on MS-COCO for a better view.}
\label{fig:visiualize}
\end{figure*}

\begin{figure*}[t!]
\centering
\includegraphics[width=1.0\textwidth]{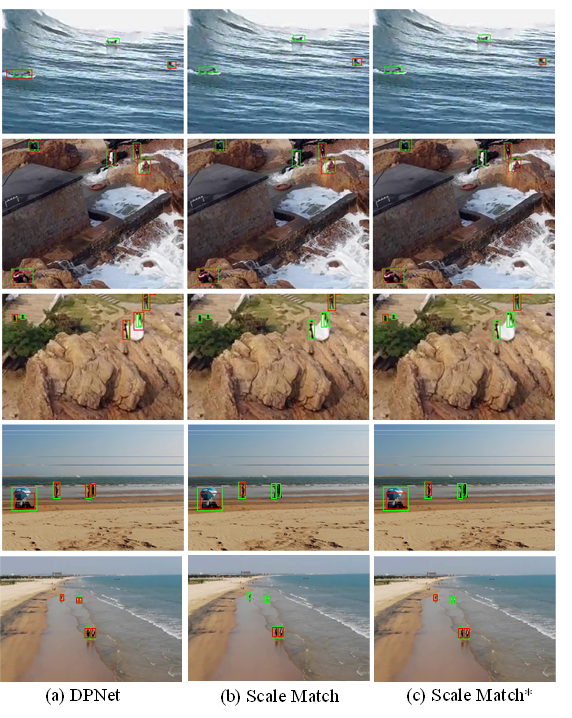}
\caption{\M{Image visualization of DPNet and other methid on TinyPerson. Scale Match and Scale Match* are RepPoints-SM and RepPoints-SM$^{\dagger}$ respectively.Green boxes denote the ground truth and red boxes denote the predicted boxes. It shows our DPNNet have less missing under same precision.}}
\label{fig:visiualize tinyperson}
\end{figure*}

\subsection{Performance Comparison}
We compare the previous network with our DPNet on TinyCOCO dataset.
As shown in Table~\ref{tab:big-table}, different backbones for our detector are used for comparison, especially some lightweight networks like ResNeXt and MobileNet. These can verify that our method is not a succedaneum of lightweight models, but is combinative with those to save computation. Lightweight models save computation mainly through reducing the complexity of the model structure. Our method reduces computation from another aspect: adaptively adjusting the size of input. The baseline means the performance is based on TinyCOCO (COCO100) and the performance of enlarging is conduct on the input resized to 800, which brings the performance gain. This shows that enlarging images can increase detection performance, especially for tiny objects. The performance in the third line represents our method, which the image are enlarged (resized up) in to 800 and dynamic down-sampling is used for detection. Our basic backbone is ResNet-50, and the performance of our method is $mAP$ 29.7. We use the larger ResNet-101 to demonstrate the consistent validity of our method, which shows that our DPNet can effectively reduce the amount of calculation.

\M{Meanwhile, we further benchmark DPNet with the state-of-the-art approaches on the TinyPerson dataset, the popular dataset for tiny object detection tasks. We have added another three representative methods in tiny object detection task,  namely RetinaNet-S-$\alpha$, RetinaNet-SM. As shown in Table~\ref{tab:big-table-tinyperson}, DPNet is almost superior to all baselines. In $AP_{tiny}$, DPNet achieves 52.33, while RetinaNet-S-$\alpha$, RetinaNet-SM are 48.34, 48.48 respectively. This indicates that our method can also achieve comparable performance with the enlarging policy under less computation cost on the TinyPerson dataset.}

Additionally, our DPNet against state-of-the-art approaches on the DOTA dataset for Horizontal box object detection in Table\ref{tab:dota}.DPNet achieves an $AP$ of 63.55, compared to 60.46 and 59.44 for faster-rcnn and repPoints, resprectively.
DPNet also achieves a SOTA efficiency-accuracy balance on the VisDrone dataset in Table\ref{tab:visdrone}, providing a superior solution for real-time small object detection in drone applications. Future work will focus on refining the dynamic strategy to further push the accuracy boundary.

\subsection{Visualization}
The prediction results of the DFP are visualized in Fig.~\ref{fig:visiualize}. The top seven images with big sizes or obvious foregrounds which occupy most part of the whole images are predicted with $0.25$ $df$. The middle seven images in which objects are smaller or lightly blurred are predicted with $0.33$ $df$. The bottom seven images have really small objects, and their hidden foregrounds nearly blend with the background; thus the largest $df$ is selected. Although the "easy" and "hard" examples may be different for humans and machines, these results are compatible with the human perception system.
\M{As shown in Fig.~\ref{fig:visiualize tinyperson}, we also compared the visualization results of DPNet, the small object domain method Scale Match, and Scale Match* using resized (up) input on the TinyPerson dataset, a real-world scenario. The results showed that DPNet had fewer missing detections at the same precision.}

\begin{figure}[t]
\centering
\includegraphics[width=.48\textwidth]{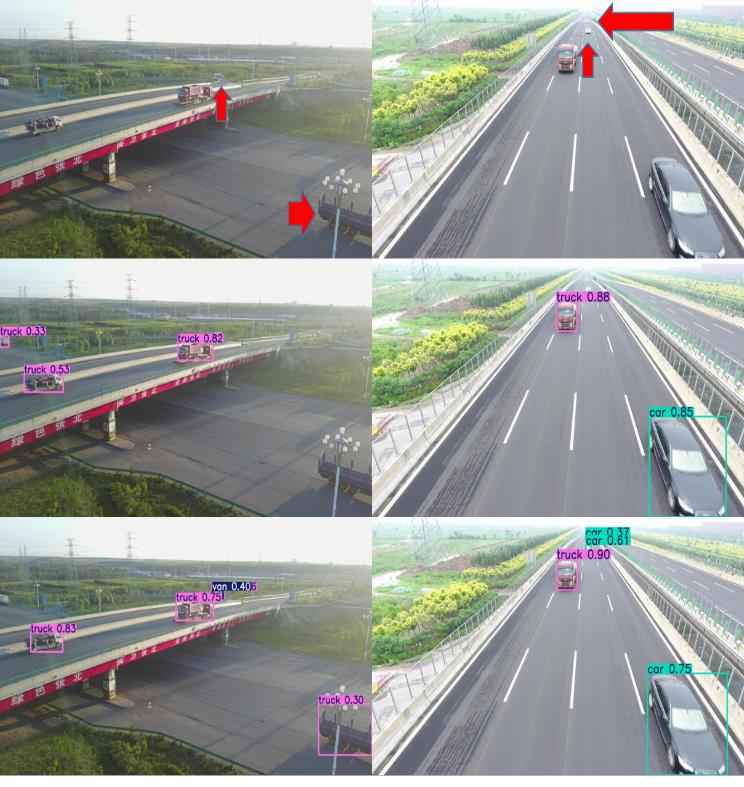}
\caption{Comparison between YOLOV11m (middle) and DPNet-MobileNet-v2 (bottom) in a simple scenario.}
\label{fig:visdrone1}
\end{figure}

We present DPNet's visualization results for small object detection in low-computing scenarios, demonstrating its advantages over the baseline YOLOv11m model. All images are sourced from the VisDrone dataset test set.
In scenarios with fewer objects, particularly for simple and common small targets, we compare the original image, YOLOv11m detection results, and DPNet-MobileNet-v2 detection results sequentially, as shown in Figure \ref{fig:visdrone1}.

On the left, YOLOv11m fails to detect a truck in the bottom-right corner, while DPNet-MobileNet-v2 successfully identifies it. In the middle, a very small target missed by YOLOv11m is detected by our method. On the right, two tiny car targets near the upper-middle region are missed by YOLOv11m but accurately detected by DPNet-MobileNet-v2, capturing both category and position.

\begin{figure}[t]
\centering
\includegraphics[width=.48\textwidth]{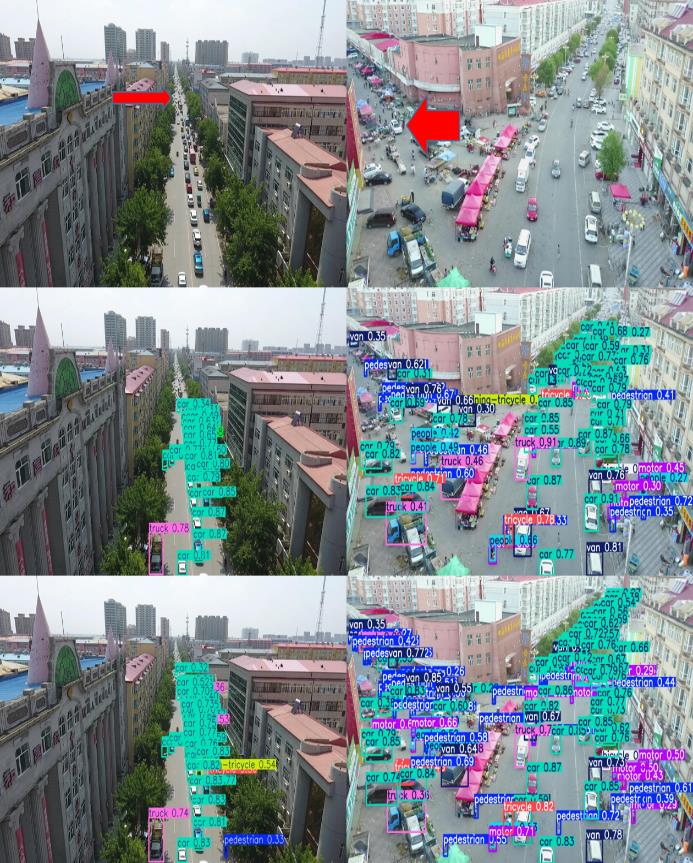}
\caption{Comparison between YOLOV11m (middle) and DPNet-MobileNet-v2 (bottom) in a dense detection scenario.}
\label{fig:visdrone0}
\end{figure}

Figure \ref{fig:visdrone0} compares the performance of YOLOv11m and DPNet-MobileNet-v2 in dense detection scenarios. On the left, YOLOv11m misses small targets at the farthest end, while DPNet mitigates this issue, though some farthest targets remain undetected. At the bottom, DPNet detects a pedestrian missed by YOLOv11m. On the right, where objects are highly concentrated, DPNet shows significantly higher detection density than YOLOv11m.

The visualization clearly demonstrates DPNet's superior performance. In comparison with the baseline method, DPNet shows greater accuracy with fewer false positives in general object detection. For planes, DPNet offers tighter bounding boxes. It also excels in detecting ships and buildings by providing more precise and correctly placed bounding boxes. Overall, DPNet delivers enhanced accuracy and precision across various object types, highlighting its effectiveness over the baseline method.
\begin{figure*}[t!]
\centering
\includegraphics[width=\textwidth]{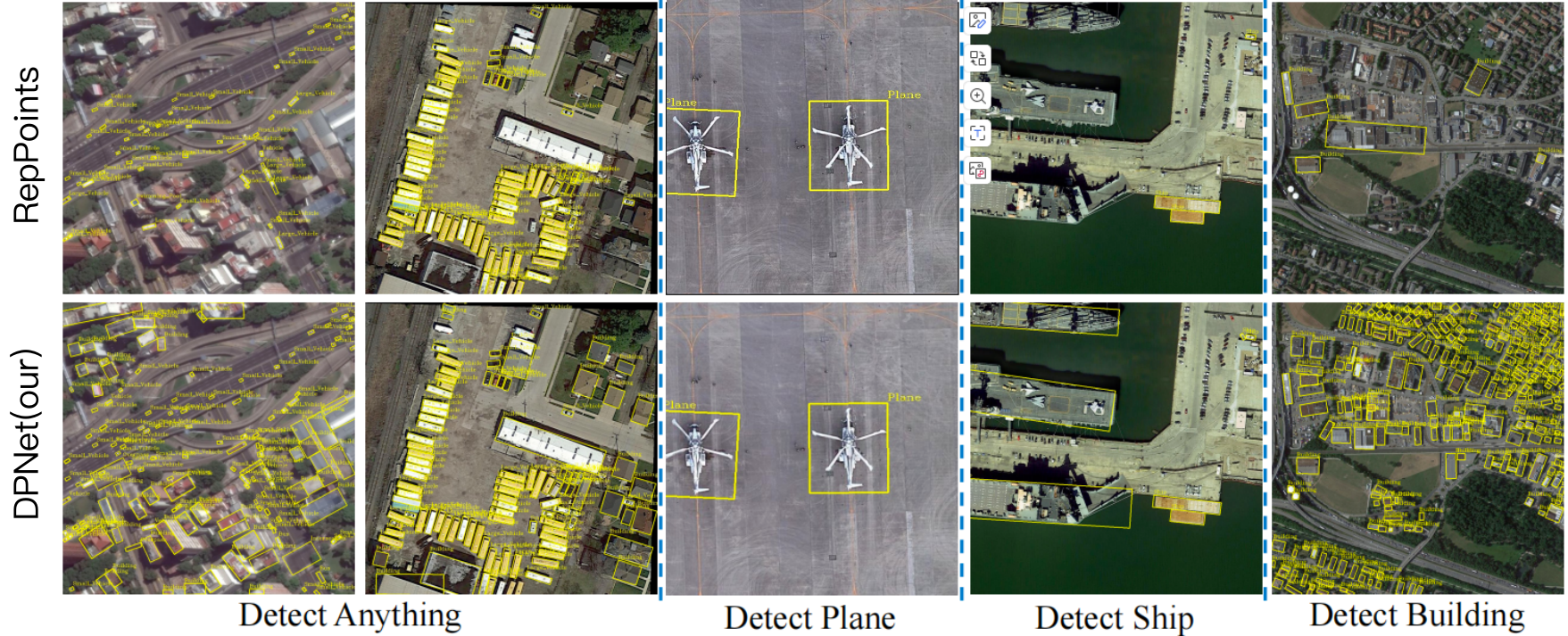}
\caption{Visualization results of RepPionts on the DOTA validation set before (the top row) and our DPNet (the bottom row), demonstrating four prompts: detecting any objects, detecting plane, detecting ship, and detecting buildings.}
\label{fig:dota}
\end{figure*}

\subsection{Analysis}
\label{subsec: analysis}
\noindent{\bfseries Why Down-Sampling Factor?} 
There are many ways to design a dynamic neural network, such as dynamic network depth, width, parameter, or feature map scale. We implement a series of experiments to simply analyze which factor is more important for the detection task. Table~\ref{tab:different depth, width and size} shows that when the depth and width of the network are changed, the performance of the detector fluctuates within a narrow range, while resizing up the input image, the performance will dramatically raises. \M{This is also why we pay attention to dynamic down-sampling.}


\section{Conclusion}
This paper gives a visual analysis of the advantage of enlarging input images to detect tiny objects. Furthermore, we propose a novel DPNet to adaptively select the most proper $df$ of a feature map. ANM is designed to solve the problem of feature aggregation inconsistency between different $df$ switches by privatizing all normalization layers for each switch in a union network.
The guidance loss is designed to better optimize the detector's performance under each $df$ in backbone through the way of re-weighting the loss of each instance in an image according to its size.
The DFP in DPNet predicts the probability distribution of $df$s, helping to choose the performance-sufficient and cost-efficient $df$ for each image. Thus, it can significantly reduce computational costs and enhance the application of recognition in unmanned aerial systems. As the first dynamic neural network suitable for detection tasks, DPNet can provide inspiration for more researchers in this field.

\section*{Acknowledgment}

This work 
The authors would like to thank the reviewers and associate editor for their valuable feedback on the paper. The authors extend their appreciation to Researcher Supporting Project National Natural Science Foundation of China (Grant No.62472043, U21A20468), Beijing Natural Science Foundation (Grant No.4232050).


%





\ifCLASSOPTIONcaptionsoff
  \newpage
\fi





\bibliographystyle{IEEEtran}
\bibliography{IEEEabrv,Bibliography}
%

\begin{IEEEbiography}[{\includegraphics[width=1in,height=1.25in,clip,keepaspectratio]{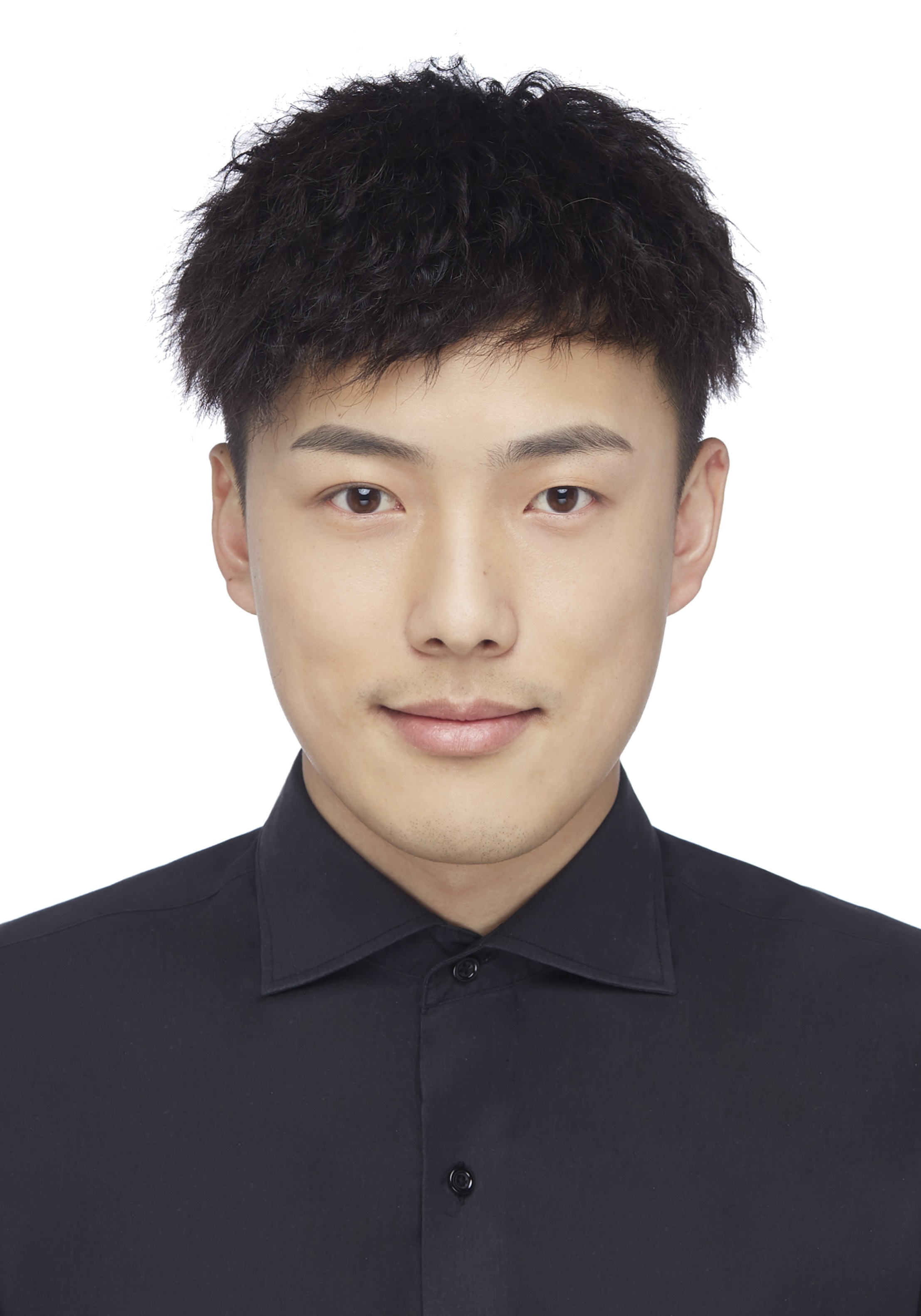}}]{Luqi Gong (Student member, IEEE) received his B.Eng. and M.Eng. degree from Shandong University of Science \& Technology, Qingdao, China in 2017 and Institute of Computing Technology, Chinese Academy of Sciences, Beijing, China in 2020, respectively. He is currently purchasing the Ph.D. degree in Beijing University Of Posts and Telecommunications. His current research interests include satellite computing, acceleration of deep neural network and embedded intelligent computing system.}
\end{IEEEbiography}
\begin{IEEEbiography}[{\includegraphics[width=1in,height=1.25in,clip]{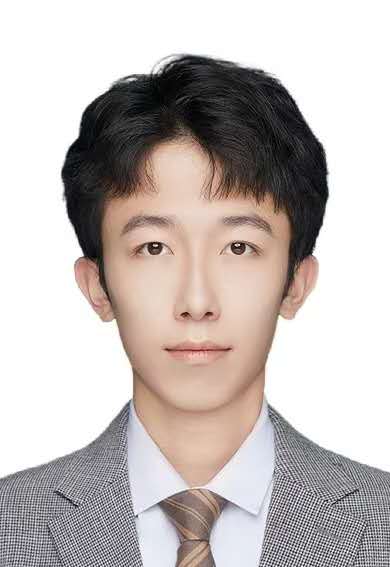}}]{Haotian Chen is currently pursuing a Bachelor's degree in Computer Science and Technology at Southwest Jiaotong University, Chengdu, Sichuan, China. With a strong academic focus on computer vision, his research interests include semantic segmentation and action recognition. In addition to his academic achievements, Haotian has demonstrated exceptional leadership and technical expertise as a project leader and technical director. His innovative contributions have earned him several prestigious accolades, including the National First Prize in the National Creative Crane Competition for Mechanical Innovation for College Students and the National Third Prize in the Challenge Cup National College Students' Innovation and Entrepreneurship Competition.} 
\end{IEEEbiography}
\begin{IEEEbiography}[{\includegraphics[width=1in,height=1.25in,clip,keepaspectratio]{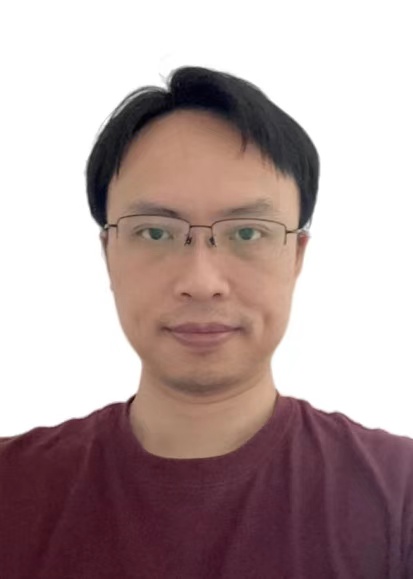}}]{Yikun Chen (Member, IEEE) received his B.Eng. and M.Eng. degree from South China Normal University, Guangzhou, China in 2011 and North Borneo University, Malaysia in 2021, respectively. His research interests encompass the application of computer vision technology in the intelligent transformation of construction sites and smart parks. He was honored with the Guangdong Province Zhuhai City Software Innovation Talent Award in 2014. }
\end{IEEEbiography}
\begin{IEEEbiography}[{\includegraphics[width=1in,height=1.25in,clip]{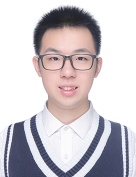}}]{Tianliang Yao (Student Member, IEEE) is currently pursuing a B.Eng. degree in automation at Tongji University, Shanghai, China. His current research interests include computer vision, medical image analysis and embodied intelligence in medical robotics. He was a recipient of the Best Conference Paper Finalist Award at the IEEE International Conference on Advanced Robotics and Mechatronics (ARM) in 2023 and IEEE ICRA RAS Travel Grant Award in 2025.}
\end{IEEEbiography}
\begin{IEEEbiography}[{\includegraphics[width=1in,height=1.25in,clip,keepaspectratio]{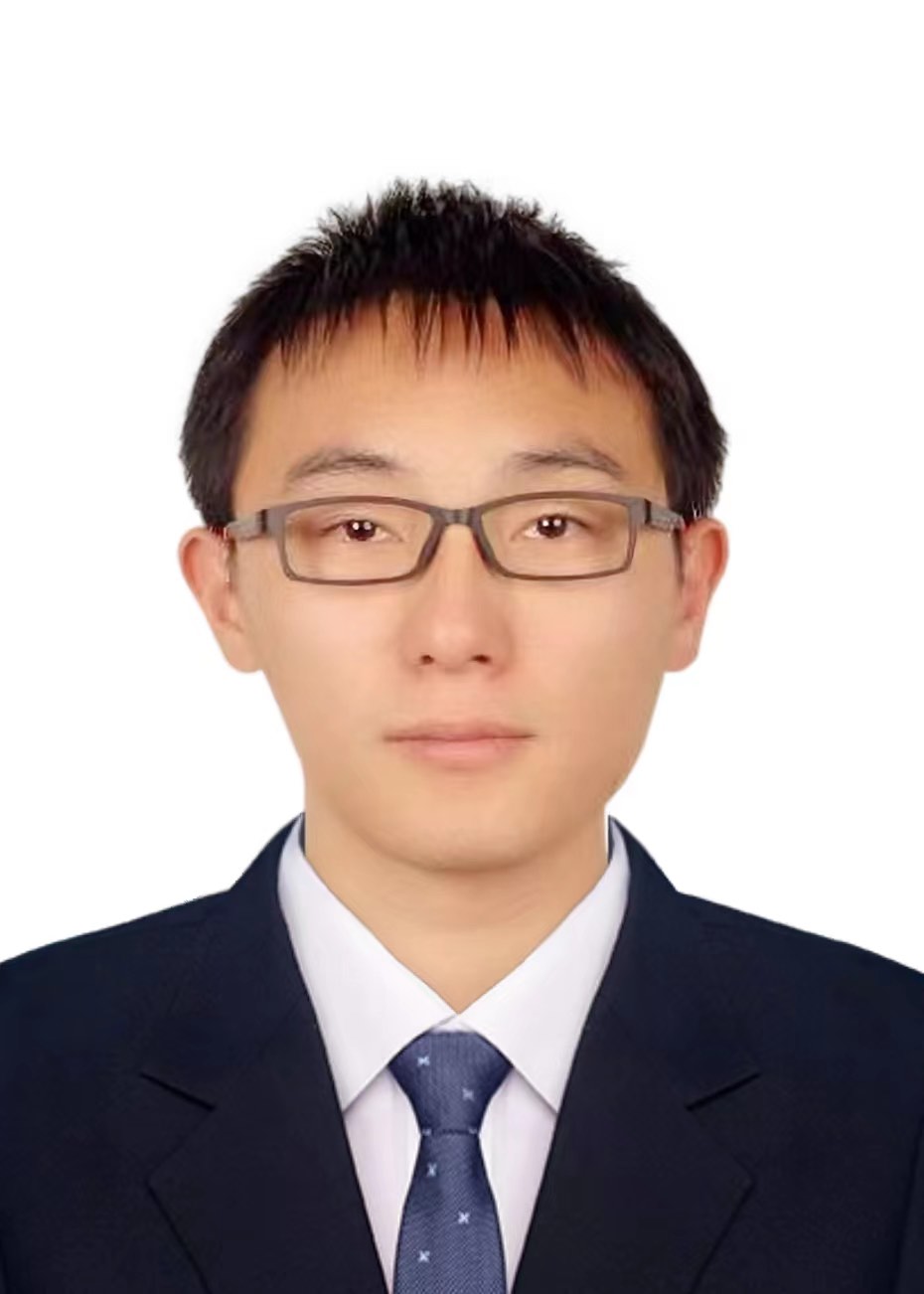}}]{Chao Li (Member, IEEE) is a principal investigator at Zhejiang Lab, Hangzhou, China, and an associate professor at the Institute of Computing Technology, Chinese Academy of Sciences, Beijing, China. His research interests include interpretable learning, machine learning, data mining, and edge intelligence systems. He has published more than 20 papers in the prestigious refereed journals and conference proceedings, such as AAAI, ICCV, ICML etc.}
\end{IEEEbiography}
\begin{IEEEbiography}[{\includegraphics[width=1in,height=1.25in,clip,keepaspectratio]{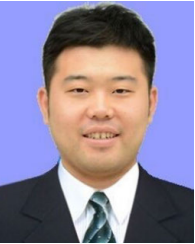}}]{Shuai Zhao (Member, IEEE) received the Ph.D. degree in computer science and technology from the Beijing University of Posts and Telecommunications in 2014. He is currently a Professor with the State Key Laboratory of Networking and Switching Technology, Beijing University of Posts and Telecommunications. His current research interests include the Internet of Things Technology, distributed learning, and intelligent systems.}

\end{IEEEbiography}
\begin{IEEEbiography}[{\includegraphics[width=1in,height=1.25in,clip,keepaspectratio]{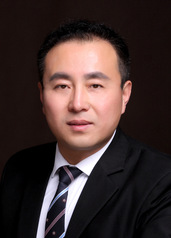}}]{Guangjie Han (Fellow, IEEE) is currently a Professor with the Department of Internet of Things Engineering, Hohai University, Changzhou, China. He received his Ph.D. degree from Northeastern University, Shenyang, China, in 2004. In February 2008, he finished his work as a Postdoctoral Researcher with the Department of Computer Science, Chonnam National University, Gwangju, Korea. From October 2010 to October 2011, he was a Visiting Research Scholar with Osaka University, Suita, Japan. From January 2017 to February 2017, he was a Visiting Professor with City University of Hong Kong, China. From July 2017 to July 2020, he was a Distinguished Professor with Dalian University of Technology, China. His current research interests include Internet of Things, Industrial Internet, Machine Learning and Artificial Intelligence, Mobile Computing, Security and Privacy. Dr. Han has over 500 peer-reviewed journal and conference papers, in addition to 160 granted and pending patents. Currently, his H-index is 75 and i10-index is 341 in Google Citation (Google Scholar). The total citation count of his papers raises above 20000 times. \\
   Dr. Han is a Fellow of the UK Institution of Engineering and Technology (FIET). He has served on the Editorial Boards of up to 10 international journals, including the IEEE TII, IEEE TCCN, IEEE Systems, etc. He has guest-edited several special issues in IEEE Journals and Magazines, including the IEEE JSAC, IEEE Communications, IEEE Wireless Communications, Computer Networks, etc. Dr. Han has also served as chair of organizing and technical committees in many international conferences. He has been awarded 2020 IEEE Systems Journal Annual Best Paper Award and the 2017-2019 IEEE ACCESS Outstanding Associate Editor Award. He is a Fellow of IEEE.} 
\end{IEEEbiography}





\vfill


\end{document}